\documentclass[journal]{IEEEtran}
\IEEEoverridecommandlockouts
\usepackage{cite}
\usepackage[numbers,sort&compress]{natbib}
\usepackage{amsmath,amssymb,amsfonts, bm}
\usepackage{algorithm}
\usepackage{algorithmic}
\usepackage{subfigure}
\usepackage{graphicx}
\usepackage{caption}
\usepackage{textcomp}
\usepackage{xcolor}
\usepackage{authblk}
\usepackage[marginal]{footmisc}
\usepackage{multirow}
\usepackage{multicol}
\usepackage{bm}
\def\BibTeX{{\rm B\kern-.05em{\sc i\kern-.025em b}\kern-.08em
    T\kern-.1667em\lower.7ex\hbox{E}\kern-.125emX}}

\title{Multi-view Low-rank Preserving Embedding: A Novel Method for Multi-view Representation}

\author{Xiangzhu Meng\thanks{X. Meng and L. Feng(corresponding author, denoted by $\ast$ ) are with the School of Computer Science and Technology, Dalian University of Technology, Dalian, China(xiangzhu\_meng@mail.dlut.edu.cn; fenglin@dlut.edu.cn)}, Lin Feng\IEEEauthorrefmark{1}, Huibing Wang \thanks{H. Wang is with the College of Information Science and Technology, Dalian Maritime University, Dalian, China(huibing.wang@dlmu.edu.cn)}}

\begin{document}
\maketitle

\begin{abstract}
    In recent years, we have witnessed a surge of interest in multi-view representation learning, which is concerned with the problem of learning representations of multi-view data. When facing multiple views that are highly related but sightly different from each other, most of existing multi-view methods might fail to fully integrate multi-view information. Besides, correlations between features from multiple views always vary seriously, which makes multi-view representation challenging. Therefore, how to learn appropriate embedding from multi-view information is still an open problem but challenging. To handle this issue, this paper proposes a novel multi-view learning method, named Multi-view Low-rank Preserving Embedding (MvLPE). It integrates different views into one centroid view by minimizing the disagreement term, based on distance or similarity matrix among instances, between the centroid view and each view meanwhile maintaining low-rank reconstruction relations among samples for each view, which could make more full use of compatible and complementary information from multi-view features. Unlike existing methods with additive parameters, the proposed method could automatically allocate a suitable weight for each view in multi-view information fusion. However, MvLPE couldn't be directly solved, which makes the proposed MvLPE difficult to obtain an analytic solution. To this end, we approximate this solution based on stationary hypothesis and normalization post-processing to efficiently obtain the optimal solution. Furthermore, an iterative alternating strategy is provided to solve this multi-view representation problem. The experiments on six benchmark datasets demonstrate that the proposed method outperforms its counterparts while achieving very competitive performance.
\end{abstract}


\section{Introduction}
In general, one object could be characterized by different kinds of views \cite{li2018survey, meng2019,xu2013survey}, because data is often collected from diverse domains or obtained from different feature extractors. For examples, web pages could be usually presented by the page-text and hyperlink information; Color, text or shape information could be used as different kinds of features, in image and video processing, such as HSV, Local Binary Pattern (LBP) \cite{ojala2002multiresolution}, Gist \cite{douze2009evaluation}, Histogram of Gradients (HoG) \cite{dalal2005histograms}, Edge Direction Histogram (EDH) \cite{gao2008image}. Since different views describe distinct properties of the instance, multiple views contain more complete information than just one view. Generally, it could achieve better performance in many real-world applications \cite{smeulders2000content, datta2008image, jiang2011fuzzy, wei2019text, tao2012discriminative, qiao2010sparsity} by taking the complementary information from multiple views into consideration. As we all know, the performance of machine learning methods is heavily dependent on the expressive power of feature representation. Consequently, multi-view representation learning has received great research efforts, such as multi-view information fusion, multi-view representation alignment, and so on. Those multi-view methods focused on exploiting the diverse information and complementary information among multiple views to achieve a comprehensive representation of the instance.

Nowadays, multi-view representation methods \cite{xia2010multiview, tang2009clustering, tzortzis2012kernel, cao2013robust, liu2014multiview, ngiam2011multimodal, su2015multi, hardoon2004canonical, rupnik2010multi, kan2016multi, andrew2013deep, zhang2018generalized} have been well studied in many applications. Multi-view information fusion methods \cite{xia2010multiview, tang2009clustering, tzortzis2012kernel, cao2013robust, liu2014multiview, ngiam2011multimodal, su2015multi} aimed to fuse multi-view features into single compact representation. Multiview Spectral Embedding (MSE) \cite{xia2010multiview} was an extension of Laplacian Eigenmaps (LE) \cite{belkin2002laplacian} and incorporated it with multi-view data to find a common low-dimensional subspace, which exploited low-dimensional representations based on the graph. Tang et al. \cite{tang2009clustering} fused the information from multiple graphs with linked matrix factorization, where each graph was approximated by the graph-specific factor and the common factor. Tzortzis et al. \cite{tzortzis2012kernel} expressed each view as a given kernel matrix and learned a weighted combination of those kernels in parallel. Multi-view sparse coding \cite{cao2013robust, liu2014multiview} associated the shared latent representation for the multi-view data by a set of linear mappings that are defined as dictionaries. Ngiam et al. \cite{ngiam2011multimodal} proposed a novel method to extract shared representations via training deep auto-encoder \cite{bengio2013representation}, which utilized the concatenation of the final hidden coding of audio and video modalities as inputs and mapped these inputs to a shared representation layer. Inspired by the great success of Convolutional Neural Networks (CNN) \cite{krizhevsky2012imagenet}, Su et al. \cite{su2015multi} introduced a multi-view CNN for 3D object recognition, which integrated information from multiple 2D views of an object into a single and compact representation. However, those multi-view fusion methods might ignore the consistent correlation information among multiple views so that compatible and complementary information couldn't be made full advantage. To capture the relationships among different views, multi-view representation alignment methods \cite{hardoon2004canonical, rupnik2010multi, kan2016multi, andrew2013deep, zhang2018generalized} were proposed to explore consistent correlation information by feature alignment. In particular, Canonical Correlation Analysis (CCA) \cite{hardoon2004canonical} and its kernel extension \cite{bach2002kernel} were representative features alignment methods, which could project two views into the common subspace by maximizing the cross correlation between two views. Furthermore, CCA was further generalized for a multi-view scenario termed as multi-view canonical correlation analysis (MCCA) \cite{rupnik2010multi}. Kan et al. \cite{kan2016multi} proposed multi-View Discriminant Analysis to extend Linear Discriminant Analysis(LDA) \cite{mika1999fisher,yu2017robust} 
based on CCA into a multi-view setting, which projected multi-view features to one discriminative common subspace. Inspired by the success of deep neural networks \cite{vincent2008extracting, bengio2013representation}, Andrew et al. \cite{andrew2013deep} proposed the method of deep CAA to capture the high-level association between multi-view data by coupling the joint representation among multiple views at the higher level. Zhang et al. \cite{zhang2018generalized} proposed a Generalized Latent Multi-View Subspace Clustering, which jointly learns the latent representation and multi-view subspace representation within the unified framework. Nevertheless, these alignment methods mainly employed the linear projection to model the cross correlation for forcible alignment of pairwise views so that algorithm performance would be not enough robust when facing such multiple views that were highly related but sometimes different from each other. 

It's also attracted wide attention to achieve the multi-view clustering agreement \cite{kumar2011co, wang2015robust, wang2016iterative, zhang2016flexible, wang2018multiview} to yield a substantial superior clustering performance over the single view paradigm. For example, Kumar et al. \cite{kumar2011co} proposed a co-regularized multi-view spectral clustering framework that captured complementary information among different viewpoints by co-regularizing the clustering hypotheses. Besides, other works in \cite{wang2017unsupervised, wang2017effective, wu2019cycle, wang2018beyond, feng2020multi, meng2020similarity} could also obtain promising performance in the multi-view learning environment. Even though the above multi-view methods have achieved promising performance in many applications, most of them could not make use of compatible and complementary information among multiple views or introduce additional learnable parameters in fusing multi-view information. Moreover, the limitations of their generalization and scalability exist all the time.

\subsection{Contributions}
In this paper, we first propose a novel single-view representation method called Low-rank Preserving Embedding (LPE), which provides with three different manners, including direct embedding, linear projection, and kernel method, to maintain the low-rank reconstruction relationships among samples. In this way, we could flexibly choose the embedding manner to preserve the low-rank reconstruction structure under each view when fusing multi-view information. Then we extend LPE into multi-view setting to develop a multi-view method called Multi-view Low-rank Preserving Embedding (MvLPE), which integrates all views into one centroid view by minimizing the disagreement between centroid view and other views and combines it with the low-rank reconstruction structure in each view. Specially, the proposed multi-view method could learn an optimal weight for each view without additive parameters when fusing all views into centroid view, and the obtained the embedding of centroid view could affect the solution of the embeddings of other views in turn. Consequently, both compatible and complementary information from multi-view feature sets and the low-rank reconstruction structure under all instances in each view could be considered at the same time. To obtain the optimal solution, we further design an iterative alternating strategy for the proposed MvLPE and also analyze its convergence. Furthermore, we discuss potential extensions for single-view methods to improve the generalization of our method. Finally, extensive experiments on six benchmark datasets demonstrate that the proposed MvLPE achieves comparable performance. The major contributions of this paper are summarized as follows:
\begin{itemize}
    \item We propose a novel single-view representation method providing with three different manners to maintain the low-rank reconstruction relationships among samples, called Low-rank Preserving Embedding (LPE). It could flexibly choose the embedding manner to keep the low-rank reconstruction structure when fusing multi-view information.
    
    \item We extend LPE into multi-view setting to propose a novel multi-view method, called Multi-view Low-rank Preserving Embedding (MvLPE), to integrate different information into one centroid view. It considers both compatible and complementary information from multi-view feature sets and the low-rank reconstruction structure under all instances in each view at the same time. 
    
    \item An effective and robust iterative alternating algorithm is developed to seek an approximate optimal solution for MvLPE. Moreover, we provide with the convergence analysis of this method and its extensions for those single-view methods.
    
    \item The experimental results on 6 benchmark datasets demonstrate that the proposed method outperforms its counterparts and achieves comparable performance.
\end{itemize}

\subsection{Organization}
The rest of the paper is organized as follows: in Section \uppercase\expandafter{\romannumeral2}, we provide briefly some related methods which have attracted extensive attention; in Section \uppercase\expandafter{\romannumeral3}, we describe the construction procedure of MvLPE and optimization algorithm for MvLPE; in Section \uppercase\expandafter{\romannumeral4}, extensive experiments on text and image datasets demonstrate the effectiveness of our proposed approach; in Section \uppercase\expandafter{\romannumeral5}, we finally conclude this paper.

\begin{figure*}[htbp]
\centering
\includegraphics[width=0.95\textwidth]{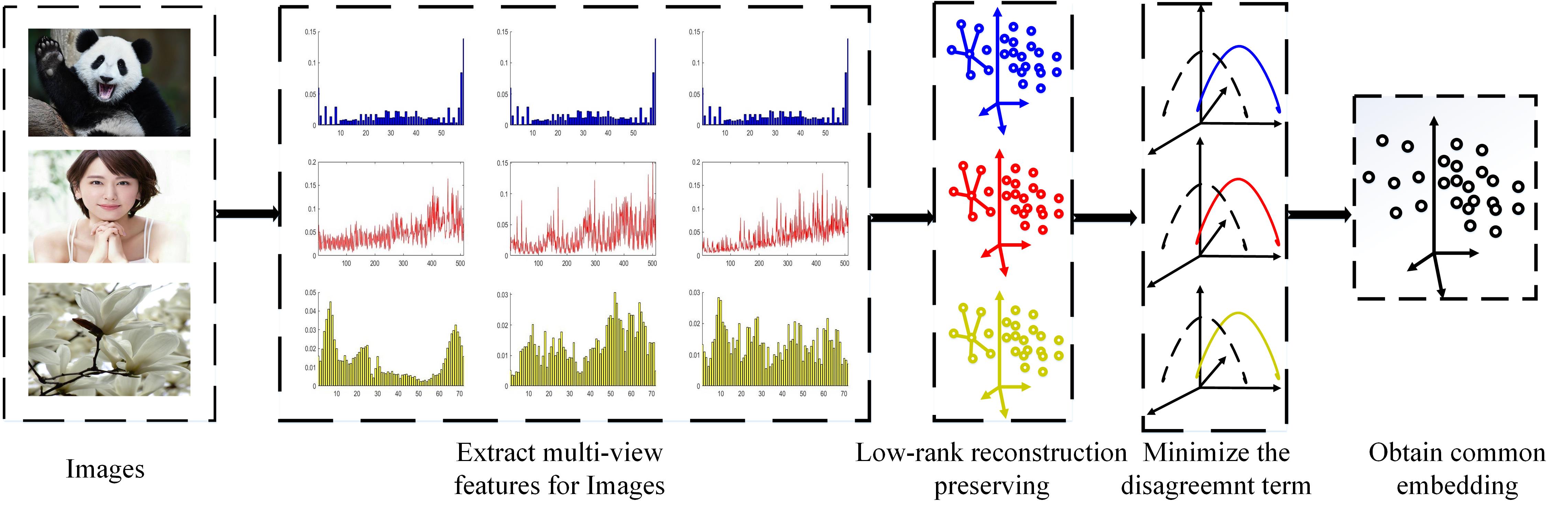}
\caption{The working procedure of Multi-view Low-rank Preserving Embedding (MvLPE), which aims to handle multi-view representation problem. Taking the images as an example, we first extract LBP, GIST, and EDH descriptors for images as multi-view information. Then, we apply the representation method called Low-rank Preserving Embedding (LPE) for images to obtain low-rank reconstruction structure and their embeddings. By minimizing the disagreement term between centroid view and all views, we integrate compatible and complementary information from multi-view feature sets to construct common embedding as multi-view representation. Finally, an iterative alternating strategy is adopted to find the optimal solution for MvLPE.}
\label{test}
\end{figure*}

\section{Related Works}
In this section, we first introduce a low-rank representation method \cite{liu2013robust}, which seeks the low-rank representation among all the candidates that can represent the data samples as linear combinations of the bases in a given dictionary. Then, we review a multi-view learning method called Multi-view Spectral Embedding \cite{xia2010multiview}.
\subsection{Low-Rank Representation}\label{LRR}
Liu et al. \cite{liu2013robust} proposed a representative low-rank representation method to handle the subspace recovery problem, which was quite superior in terms of its effectiveness, intuitiveness and robustness to noise corruptions. Assume that we are provided a features set consisting of $N$ samples, which are extracted from the $v$th view. We denote the features set in the $v$th view as $\bm{X}^v=[{\bm{x}_1^v}, {\bm{x}_2^v}, \ldots , {\bm{x}_N^v}]$. When we choose the matrix $\bm{X}^v$ itself as a dictionary that linearly spans the data space. We could get the following optimization problem:
 \begin{equation}
 \begin{split}
 &\mathop {\min }\limits_{\bm{Z}^v,\bm{E}^v} { rank(\bm{Z}^{v}) + }\lambda {\left\| \bm{E}^v \right\|_{2,1}} \\
 & s.t.\hspace{0.5em}{ \bm{X}^{v} = \bm{X}^{v}\bm{Z}^{v} + \bm{E}^{v}}\\
 \end{split}
 \end{equation}
 where $\lambda$ is a hyperparameter and $\bm{Z}^v \in {\mathbb{R}^{N \times N}}$ is the lowest rank representation of data $\bm{X}^v$. After obtaining an optimal solution, we could recover the original data by using $\bm{X}^{v}\bm{Z}^{v}$. Since rank($\bm{X}^{v}\bm{Z}^{v}$)$\leq$rank($\bm{X}^{v}$), $\bm{X}^{v}\bm{Z}^{v}$ is also a low-rank recovery to the original data. By choosing an appropriate dictionary, as we will see, the lowest-rank representation can recover the underlying row space so as to reveal the true segmentation of data. Therefore, LRR could handle well the data drawn from a union of multiple subspaces.

\subsection{Multi-view Spectral Embedding}
Multi-view Spectral Embedding (MSE) \cite{xia2010multiview} is a spectral embedding method that could map different features from multiple views into a common subspace. It aims to find a low-dimensional and physically meaningful embedding. Assume that given data has m views. Let ${\bm{X}^v}=\left\{ {\bm{x}_1^v,\bm{x}_{_{}^2}^v, \ldots ,\bm{x}_N^v} \right\}$ denote the features set in the $v$th view. MSE aims to find a low-dimensional embedding $\bm{U}$ as follows:
\begin{equation}
\begin{split}
&\mathop {\max }\limits_{\bm{U}, {\bm{\alpha}}^v \in {\mathbb{R}^{N \times k}}} \sum\limits_{v = 1}^m {\bm{\alpha}}^v {tr({\bm{U}}^{^T}{\bm{L}^v}{\bm{U}})}  \\
&\hspace{1em}s.t.\hspace{1.5em}{\bm{U}}^{^T}{\bm{U}}{ =  \bm{I},} \sum\limits_{v = 1}^m {{\alpha ^v}}  = 1,\forall {\rm{  1}} \le v \le m{\rm{    }}\\
\end{split}
\end{equation}
where ${\bm{L}^v}$ denotes the normalized graph Laplacian matrix in the $v$th view, $\alpha  = \left[ {{\bm{\alpha} ^1},{\bm{\alpha} ^2}, \cdots ,{\bm{\alpha} ^m}} \right]$ is a non-negative weight vector. And ${\bm{\alpha} ^v}$ reflects the importance which the view ${\bm{X}^v}$ plays in learning to obtain the low-dimensional embedding. Global coordinate alignment is utilized such that low-dimensional embedding in different views could keep consistent with each other globally. And, to solve the above problem, an iterative method could be adopted to update $\bm{\alpha}$ and $\bm{U}$ respectively.

\begin{table}[htbp]
\caption{Important notations used in this paper}
\label{notations}
\centering
\begin{tabular*}{0.45\textwidth}{@{\extracolsep{\fill}}cl}  
\hline
Notaiton & Description\\
\hline  
$X^v$ & The set of all instances in the $v$th view \\
$x_i^v$ & The $i$th instance in the $v$th view \\
$d^v$ & Dimension of the subspace in the the $v$-th view \\
$Z^v$ & The low-rank reconstruction weights in the $v$th view \\
$M^v$ & The low-rank reconstructive matrix for the the $v$-th view \\
$\bm{w}^v$ & The projection direction for the the $v$-th view \\
$K^v$ & The kernel matrix for the the $v$-th view \\
$\bm{\psi}(\bm{X}^v)$ & The set of all low-deimensional embedding in the $v$th view \\
$\bm{w}^v$ & The weight coefficient for the the $v$-th view \\
$U^*$ & The low-dimensional embedding for the centroid view \\
$\bm{1}_{1/N}$ & The ${N \times N}$ matrix that each element is filled with $1/N$\\
$\bm{I}_N$ & The ${N \times N}$ identity matrix\\
$\bm{I}_{d^v}$ & The ${d^v \times d^v}$ identity matrix\\
m & The number of views \\
N & The number of samples \\
\hline
\end{tabular*}
\end{table}

\section{Multi-view Low-Rank Preserving Embedding}
In this section, we first propose a new single-view representation method called Low-rank Preserving Embedding (LPE), which provides with three different manners to maintain the low-rank reconstruction relationship among samples. Then, we extend LPE into the multi-view setting to propose a multi-view method called Multi-view Low-rank Preserving Embedding (MvLPE), which fully integrates compatible and complementary information from multi-view features sets to construct common embedding for all views. Then, an iterative alternating strategy is derived to find the optimal solution for our method and the optimization procedure is illustrated in detail. Fig.\ref{test} shows the working procedure of MvLPE. Moreover, we provide the convergence discussion of our method in detail and the extensions for those single-view methods according to our proposed MvLPE. For convenience, the important notations in this paper are listed in Table \ref{notations}.

\subsection{Low-rank Preserving Embedding}
Recall that we are provided a features set consisting of $N$ samples, which is extracted from the $v$th view. We denote the features set in the $v$th view as $\bm{X}^v=[{\bm{x}_1^v}, {\bm{x}_2^v},  \ldots , {\bm{x}_N^v}]$. As discussed in Section \ref{LRR}, LRR employs $\bm{X}^v$ itself as a dictionary, which exists such two issues consisting of unsuitable correlation reconstruction and high computational cost caused by the number of instances. Inspired by the fact that the compact combination of samples always lies on the local subspace of the test sample, we replace dictionary $\bm{X}^v$ with the near neighbors set responding to each sample. As a result, we could achieve more ideal space reconstruction than LRR. Meanwhile, the issue of high computational cost could be handled by choosing its near neighbors for individual sample, which could significantly reduce the scale of the dictionary. Therefore, it's feasible and necessary to choose such a dictionary for individual sample by using $K$ its near neighbors. Combining this with the low-rank hypothesis, we could get the following optimization problem:
 \begin{equation}
 \begin{split}
 & \mathop {\min }\limits_{\bm{Z}^v,\bm{E}^v} {rank(\bm{Z}^v) + }\lambda {\left\| \bm{E}^v \right\|_F^2} \\
 & s.t.\hspace{0.5em}{\bm{X}_i^v = {\widetilde{\bm{X}}_i^v \bm{Z}_i^v} + {\bm{E}_i^v}, {\forall 1 \le i \le N}} \\
 \end{split}
 \end{equation}
where $\widetilde{\bm{X}}_i^v$ is the dictionary of $i$th sample consisting of $K$ its closed neighbors, $\bm{Z}_i^v$ and $\bm{E}_i^v$ denote the $i$th column data in the matrix $\bm{Z}^v$ and $\bm{E}^v$ respectively. 
As a common practice in rank minimization problems, we replace the rank function with the nuclear norm and subject to the constraints the columns of the matrix $\bm{Z}_i^v$ sum to one. By this means, it is deduced to the following optimization problem:
  \begin{equation}\label{low_rank_coef}
  \begin{split}
  &\mathop {\min }\limits_{\bm{Z}^v,\bm{E}^v} {\| \bm{Z}^v \|_* + }\lambda {\| \bm{E}^v \|_F^2} \\
  &{ s}{.t.}\hspace{0.5em}{ \bm{X}_i^v = \widetilde{\bm{X}}_i^v\bm{Z}_i^v + \bm{E}_i^v, {\bm{Z}_i^v}^T\bm{1}=1, \forall 1 \le i \le N}\\
 \end{split}
 \end{equation}
 
And we aim to maintain the low-rank reconstruction relationships among samples, which are obtained by Eq.(\ref{low_rank_coef}). For the convenience of modeling and solving, a simple trick is used to transform the matrix $\bm{Z}^v \in {\mathbb{R}^{K \times N}}$ into a matrix $\bm{M}^v \in {\mathbb{R}^{N \times N}}$, which fills column elements in the matrix $\bm{M}^v$ according to the low-rank coefficients $\bm{Z}_i^v$ of its neighbors and fills zeros into other elements. Accordingly, we could define the following objective function to seek low dimensional embedding while maintaining the low-rank reconstruction relationships:
\begin{equation}\label{lrp_embedding}
\begin{split}
&\mathop {\min }\limits_{\bm{U}^v} \hspace{0.5em}tr\left( {\bm{U}^v{{({\bm{I}_N} - \bm{M}^v)}^T}({\bm{I}_N} - \bm{M}^v){{\bm{U}^v}^T}} \right) \\
&\hspace{0.5em} s.t. \hspace{1em}\bm{U}^v{{\bm{U}^v}^T} = {\bm{I}_{d^v}} \\
\end{split}
\end{equation} 
where $\bm{U}^v\in {\mathbb{R}^{d^v \times N}}$ denotes the embedding in the $v$th view, $d^v$ is the dimension of $\bm{U}^v$, and $tr(\cdot)$ denotes the matrix trace. Furthermore, we further propose two additional variants, which are based on linear transform and kernel trick respectively.

Suppose that $\bm{W}^v$ is a transformation matrix, that is $\bm{U}^v=\bm{W}^{v^T}\bm{X}^v$, which is a linear approximation. By simple algebra formulation, the objective function in Eq.(\ref{lrp_embedding}) can be expressed as follows:
\begin{equation}\label{lrp_linear}
\begin{split}
&\mathop {\min }\limits_{\bm{W}^v} \hspace{0.5em}tr\left( {\bm{W}^{v^T}\bm{X}^v{{({\bm{I}_N} - \bm{M}^v)}^T}({\bm{I}_N} - \bm{M}^v)}{\bm{X}^v}^T{\bm{W}^v} \right) \\
&\hspace{0.5em} s.t. \hspace{1em}\bm{W}^{v^T}\bm{X}^v{\bm{X}^v}^T{\bm{W}^v} = {\bm{I}_{d^v}} \\
\end{split}
\end{equation}

Furthermore, suppose that the Euclidean space is mapped to a Hilbert space, that is $\bm{X}_{\phi}^v=[\phi^v(\bm{x}_1^v), \phi^v(\bm{x}_2^v), \ldots, \phi^v(\bm{x}_N^v)]$, where $\phi^v(\cdot)$ is a nonlinear map. It has been verified \cite{scholkopf1997kernel} that $\bm{W}_{\phi}^v$ is that mapped space spanned by $\phi^v(\bm{x}_1^v), \phi^v(\bm{x}_2^v), \ldots, \phi^v(\bm{x}_N^v)$. Consequently, $\bm{W}_{\phi}^v$ could be expressed as follows:
\begin{equation}
    \begin{split}
        \bm{W}_{\phi}^v=\sum\limits_{i = 1}^N {\phi^v(\bm{x}_i^v)\bm{\beta}_i^v}=\bm{X}_{\phi}^v\bm{\beta}^v
    \end{split}
\end{equation}
 where $\bm{\beta}^v={[\bm{\beta}_1^v, \bm{\beta}_2^v, \ldots, \bm{\beta}_N^v]}^T \in \mathbb{R}^{N \times d^v}$ consists of the expansion coefficients. Set $\bm{K}_{\phi}^v = {\bm{X}_{\phi}^v}^T \bm{X}_{\phi}^v$. Combining this with the Eq.(\ref{lrp_linear}), we could obtain the following low-rank preserving problem based on kernel:
 \begin{equation}\label{lrp_kernel}
\begin{split}
&\mathop {\min }\limits_{\bm{\beta}^v} \hspace{0.5em}tr\left( {\bm{\beta}^{v^T}\bm{K}_\phi^v{{({\bm{I}_N} - \bm{M}^v)}^T}({\bm{I}_N} - \bm{M}^v)}\bm{K}_\phi^v \bm{\beta}^v \right) \\
&\hspace{0.5em} s.t. \hspace{1em}\bm{\beta}^{v^T} \bm{K}_{\phi}^v \bm{K}_{\phi}^v \bm{\beta}^v = {\bm{I}_{d^v}} \\
\end{split}
\end{equation}

In terms of the discussion above, we provide with three manners to obtain the low-dimensional embedding based on preserving the low-rank reconstruction relation among the samples. Therefore, we could obtain a unified low-rank preserving embedding method, including direct embedding, linear transform, and kernel method, so that we could more flexibly choose embedding manner to fully preserve the low-rank reconstruction relations information among samples when fusing multi-view features based on LPE. For convenience, we utilize $\bm{\psi}(\bm{X}^v)$ to generally stand for the low dimensional embedding obtained by low-rank preserving embedding method. That is $\bm{\psi}(\bm{X}^v)=\bm{U}^v$, $\bm{W}^{v^T}\bm{X}^v$, or $\bm{\beta}^{v^T} \bm{K}_{\phi}^v$, which is responding to different modes of low-rank preserving embedding respectively.

\subsection{The construction process of Multi-view Low-rank Preserving Embedding}
When facing with multi-view problems, solving the problem for all views separately will fail to integrate multi-view features and make favorable use of the complementary information from multiple views. For solving this problem, we propose a multi-view method called Multi-view Low-rank Preserving Embedding (MvLPE), to fully apply all features from different views into one centroid view and learn common representations, which extends LPE into the multi-view setting. However, the dimension of the features set in each view owns its size, which is different from the other views. Besides, it isn't easy to obtain common embedding directly because of its intrinsic geometric properties in each view. Therefore, integrating different views into one centroid view is still full of challenges.

Inspired by these works \cite{kumar2011co, meng2020similarity}, we firstly make such hypothesis that similarities among the instances in each view and the centroid view should be consistent under the novel representations. This hypothesis means that all similarity matrices from the $v$th views should be consistent with the similarity of the centroid view by aligning the similarities matrix computed from the centroid view and the $v$th view. Noting that, $\bm{U}^*$ in the centroid view and $\bm{\psi}(\bm{X}^v)$ in the $v$th view have different dimensions $d^*$ and $d^v$. To implement this hypotheses and deal with the dimensional difference problem, we utilize the following cost function as a measurement of agreement between the centroid view and the $v$th view:
\begin{equation}\label{reg0}
\begin{split}
&Sim\left( {{\bm{U}^*},{\bm{\psi}(\bm{X}^v)}} \right) = -\left\| {{\bm{K}^*} - {\bm{K}^v}} \right\|_F^p\\
\end{split}
\end{equation}
where $\bm{K}^*$ and $\bm{K}^v$ stand for the similarity matrix of the centroid view and the $v$th view respectively, $\left\|  \cdot  \right\|_F^p$ denotes the exponential function of Frobenius norm ($F$-norm), and $0 < p \le 2$ is a scalar. With the change of the value of $p$, a series of exponential function could be utilized. In fact, we could choose the general kernel function as our similarity measurement, such as linear kernel, polynomial kernel, Gaussian kernel and so on. For example, when we choose linear kernel as similarity measurement in the centroid view, $\bm{K}^v(\bm{U}_i^*, \bm{U}_j^*)=\bm{U}_i^{*^T}\bm{U}_j^*$ denotes the similarity between the instance $\bm{U}_i^*$ and the instance $\bm{U}_j^*$. In this way, $Sim\left( {{\bm{U}^*},{\bm{\psi}(\bm{X}^v)}} \right)$ reflects the consensus measure of the pairwise similarity among all instances under the centroid view and the $v$th view.

To further express the consensus term, we expand Eq.(\ref{reg0}) as follows:
\begin{equation}\label{reg1}
\begin{split}
&Sim\left( {{\bm{U}^*},{\bm{\psi}(\bm{X}^v)}} \right) =  -\left\| {{\bm{K}^*} - {\bm{K}^v}} \right\|_F^p\\
& = {(-tr({\bm{K}^*}^T\bm{K}^*+{\bm{K}^v}^T\bm{K}^v-{\bm{K}^*}^T\bm{K}^v-{\bm{K}^v}^T\bm{K}^*))}^{\frac{p}{2}}\\
& = {(2tr(\bm{K}^*\bm{K}^v)-tr(\bm{K}^*\bm{K}^*)-tr(\bm{K}^v\bm{K}^v))}^{\frac{p}{2}}\\
\end{split}
\end{equation}
 In the Eq.(\ref{reg1}), the second term $tr(\bm{K}^*\bm{K}^*)$ and the third term $tr(\bm{K}^v\bm{K}^v)$ in the above equation just depend on individual view, so these two terms couldn't work in integrating two different views. Consequently, we could approximate the consensus term as follows:
 \begin{equation}\label{reg2}
\begin{split}
&Sim\left( {{\bm{U}^*},{\bm{\psi}(\bm{X}^v)}} \right)  = {(tr(\bm{K}^*\bm{K}^v))}^{\frac{p}{2}}\\
\end{split}
\end{equation}
 Even though the consensus term in Eq.(\ref{reg2}) could work in integrating multi-view information, it's full of challenge to choose suitable kernel function for centroid view and each view at the same time under the consideration for the solving process and effectiveness. Besides, how to directly solve the consensus term based on such complicated kernel function is not easy. Therefore, constructing a meaningful and feasible term that reflects the consistent information between centroid view and each view is very necessary. Inspired by the above hypothesis minimizing the gap between the similarities computed in the centroid view and the similarities in the $v$th view, we expect that the distance between two instances in the centroid view is expected to be smaller if the similarity between two instances in the $v$th view is larger. In this way, we could formulate the following disagreement term by utilizing the square of Euclidean distance to substitute the matrix similarity among all instances in the centroid view:
\begin{equation}\label{reg3}
\begin{split}
&Dis\left( {{\bm{U}^*},{\bm{\psi}(\bm{X}^v)}} \right) ={\left( \sum\limits_{i,j=1}^{n}{\left\| {\bm{U}_{i}^*-\bm{U}_{j}^*} \right\|_2^2 \bm{K}_{ij}^v}\right)}^{\frac{p}{2}}\\
& ={tr\left( \bm{U}^* (\bm{D}^v-\bm{K}^v) {\bm{U}^*}^T \right) }^{\frac{p}{2}}\\
& \\
\end{split}
\end{equation}
where $\bm{D}^v=diag(\bm{d}_{11}^v, \bm{d}_{22}^v, \ldots, \bm{d}_{NN}^v)$ is a diagonal matrix, $\bm{d}_{NN}^v=\sum_{i=1}^N{\bm{K}_{iN}^v}$. Accordingly, we just consider the choice of kernel function for each view but centroid view, which is more convenient to solve.  

To integrate rich information among different features, we could obtain the following optimization problem by adding up cost function in Eq.(\ref{reg3}) among all views:
\begin{equation}\label{MvLPE_0}
\begin{split}
&\mathop {\min }\limits_{{\bm{U}^*}} {\sum\limits_{v = 1}^m {tr\left( \bm{U}^*(\bm{D}^v-\bm{K}^v) {\bm{U}^*}^T  \right)}}^{\frac{p}{2}} \\
& s.t.\hspace{0.5em}{\bm{U}^*}{\bm{U}^*}^{^T} = \bm{I}_{d^*} \\
\end{split}
\end{equation}
The Lagrange function of Eq.(\ref{MvLPE_0}) could be written as follows:
\begin{equation}\label{MvLPE_1}
\begin{split}
&{\sum\limits_{v = 1}^m {tr\left( \bm{U}^* (\bm{D}^v-\bm{K}^v)  {\bm{U}^*}^T \right)}}^{\frac{p}{2}}+\bm{\mathcal{G}}(\bm{\mathcal{A}}, \bm{U}^*)\\
\end{split}
\end{equation}
where $\bm{\mathcal{A}}$ is the Lagrange multiplier, $\bm{\mathcal{G}} (\bm{\mathcal{A}}, \bm{U}^*)$ is the formalized term derived from constraints. Taking the derivative of Eq.(\ref{MvLPE_1}) w.r.t $\bm{U}^*$ and setting the derivative to zero, we have 
\begin{equation}\label{MvLPE_2}
\begin{split}
&\sum\limits_{v = 1}^m \bm{w}^v\frac{\partial{tr\left( \bm{U}^* (\bm{D}^v-\bm{K}^v) {\bm{U}^*}^T \right)}} {\bm{U}^*}+\frac{\partial\bm{\mathcal{G}}(\bm{\mathcal{A}}, \bm{U}^*)}{\bm{U}^*}=0\\
\end{split}
\end{equation}
where 
\begin{equation}\label{solve_w}
\begin{split}
&\bm{w}^v=\frac{p}{2}{tr\left( \bm{U}^* (\bm{D}^v-\bm{K}^v)  {\bm{U}^*}^T \right)}^{\frac{p}{2}-1}\\
\end{split}
\end{equation}
It's easy to find that $\bm{w}^v>0$ is depended on the target variable $\bm{U}^*$, so Eq.(\ref{MvLPE_2}) couldn't be directly solved. If $\bm{w}^v$ is set to be stationary, Eq.(\ref{MvLPE_0}) could be considered as the solution of the following equation:
\begin{equation}\label{MvLPE_3}
\begin{split}
&\mathop {\min }\limits_{{\bm{U}^*}} \sum\limits_{v = 1}^m {\bm{w}^v tr\left( \bm{U}^* (\bm{D}^v-\bm{K}^v) {\bm{U}^*}^T  \right)} \\
& s.t.\hspace{0.5em}{\bm{U}^*}{\bm{U}^*}^{^T} = \bm{I}_{d^*} \\
\end{split}
\end{equation}
To further analyze the $\bm{w}^v$, we add the normalization on $\bm{w}^v$ in Eq.(\ref{MvLPE_3}) after calculating $\bm{w}^v$ by Eq.(\ref{solve_w}), i.e. $\sum\limits_{v = 1}^m {\bm{w}^v}=1$. If the $v$th view is close to the centroid view, then $tr\left( \bm{U}^* (\bm{D}^v-\bm{K}^v) {\bm{U}^*}^T \right)$ should be small, thus the learned weight $\bm{w}^v$ for the $v$th view is large. Accordingly, such view that isn't close to the centroid view will be assigned a small weight. Therefore, our method optimizes the weight $\bm{w}$ meaningfully. Accordingly, $\bm{w}^v$ could be realized as the weights of different views, which play different contribution in obtaining the common embedding $\bm{U}^*$. 

To further utilize low-rank reconstruction structure information in each view, we expect that the low-dimensional embedding in each view could also be adjusted by minimizing the disagreement measurement against the centroid view rather than only obtained by its low-rank structure. As a result, not only the low-rank structure in this view could be considered but complementary information from other views and centroid view would be utilized when solving the low-dimensional embedding in each view. Therefore, combining the loss function in Eq.(\ref{MvLPE_3}) with the LPE objectives across all views, we can get the joint loss function for MvLPE as follows:
\begin{equation}\label{MvLPE}
\begin{split}
&\mathop {\min }\limits_{{\bm{U}^*},\bm{\psi}(\bm{X}^1),\bm{\psi}(\bm{X}^2),\ldots, \bm{\psi}(\bm{X}^m)} \gamma \sum\limits_{v = 1}^m {\bm{w}^v tr\left( \bm{U}^* (\bm{D}^v-\bm{K}^v) {\bm{U}^*}^T  \right)} +\\
& \sum\limits_{v = 1}^m {tr\left( {{\bm{\psi}(\bm{X}^v)}{{(\bm{I}_N - {\bm{M}^v})}^T}(\bm{I}_N - {\bm{M}^v}) {\bm{\psi}(\bm{X}^v)}^{^T}} \right)}\\ 
& s.t.\hspace{0.5em}{\bm{U}^*}{\bm{U}^*}^{^T} = \bm{I}_{d^*},{\bm{\psi}(\bm{X}^v)}{\bm{\psi}(\bm{X}^v)}^{^T} = \bm{I}_{d^v},\forall 1 \le v \le m\ \\
\end{split}
\end{equation}
where $\gamma$ is a hyperparameter that controls the trade-off between the two terms of equation (\ref{MvLPE}). The first term is the agreement between the centroid and all views to follow the multi-view subspace hypotheses. The second term is the sum of LPE loss function for all views. From Eq. (\ref{MvLPE}), we could find that different embedding $\bm{\psi}(\bm{X}^v)$ inflects each other for the centroid representations. Differing from those fusion methods with additional parameters, our proposed method could automatically assign an optimal weight for each view according to theoretical explanations. Besides, the disagreement term based on distance or similarity matrix encourages to keep consistency between centroid view and other views, which is more robust and scalable than those multi-view representation methods based on features alignment. Therefore, the process of minimizing Eq. (\ref{MvLPE}) aims to find the common embedding which could integrate features from multiple views and preserve low-rank structure among instances.

\subsection{Optimization Process for MvLPE}
In this section, we provide the optimization process for MvLPE in detail. In order to find the optimal solution of Eq.(\ref{MvLPE}), we develop an algorithm based on alternative strategy, which separates the problem into several sub-problems such that each sub-problem is tractable. That is, we alternatively update each variable when fixing others. And we summarized the optimization process in Algorithm 1.

$\bm{Updating} \quad \bm{U}^*$: By fixing all variables but $\bm{U}^*$, Eq.(\ref{MvLPE}) will reduce to the following equation without considering constant additive and scaling term:
\begin{equation}
\begin{split}
&\mathop {\min }\limits_{{\bm{U}^*}} \sum\limits_{v = 1}^m {\bm{w}^v tr\left( \bm{U}^* (\bm{D}^v-\bm{K}^v) {\bm{U}^*}^T \right)} \\
& s.t.\hspace{0.5em}{\bm{U}^*}{\bm{U}^*}^{^T} = \bm{I}_{d^*} \\
\end{split}
\end{equation}
which has a feasible solution. According to the operational rules of matrix trace, the above equation could be transformed as follows:
\begin{equation}
\begin{split}
&\mathop {\min }\limits_{{\bm{U}^*}} tr\left( \bm{U}^*  \left(\sum\limits_{v = 1}^m {\bm{w}^v (\bm{D}^v-\bm{K}^v)}\right) {\bm{U}^*}^T \right) \\
& s.t.\hspace{0.5em}{\bm{U}^*}{\bm{U}^*}^{^T} = \bm{I}_{d^*} \\
\end{split}
\end{equation}
Set $L^*=\sum\limits_{v = 1}^m {\bm{w}^v (\bm{D}^v-\bm{K}^v)}$. Therefore, with the constraint ${\bm{U}^*}{\bm{U}^*}^{^T} = \bm{I}_{d^*}$, the optimal $\bm{U}^*$ could be solved by eigen-decomposition. $\bm{U}^*$ consists of eigenvectors corresponding to the smallest $d^*$ eigenvalues. 

$\bm{Updating} \quad \bm{\psi}(\bm{X}^v)$: 
By fixing all variables but $\bm{U}^v$, Eq.(\ref{MvLPE}) will reduce to the following equation :
\begin{equation}\label{solve_common_view}
\begin{split}
&\mathop {\min }\limits_{\bm{\psi}(\bm{X}^v)} {tr\left( {{\bm{\psi}(\bm{X}^v)}{{(\bm{I}_N - {\bm{M}^v})}^T}(\bm{I}_N - {\bm{M}^v}){\bm{\psi}(\bm{X}^v)}^{^T}} \right)}\\ &+\gamma {\bm{w}^v tr\left( \bm{U}^* (\bm{D}^v-\bm{K}^v) {\bm{U}^*}^T  \right)} \\
& s.t.\hspace{0.5em}{\bm{\psi}(\bm{X}^v)}{\bm{\psi}(\bm{X}^v)}^{^T} = \bm{I}_{d^v} \\
\end{split}
\end{equation}
Noting that the above equation isn't easy to be directly solved, because the expression of $K^v$ isn't readily certain and the disagreement term in Eq.(\ref{reg3}) is unsymmetric, that is $Dis(\bm{U}^*, \bm{\psi}(\bm{X}^v)) \neq Dis(\bm{\psi}(\bm{X}^v), \bm{U}^*)$. Inspired by co-training methods, which limit the search for the compatible hypothesis that predict the same labels for co-occurring in each view, we utilize the $Dis(\bm{\psi}(\bm{X}^v), \bm{U}^*)$ as the disagreement measurement between the $v$th view and the centroid view rather than $Dis(\bm{U}^*, \bm{\psi}(\bm{X}^v))$ when fixing the centroid view $\bm{U}^*$. Based on the above assumption that $\bm{w}^v$ is set to be stationary, the above equation could be further transformed as follows:
\begin{equation}\label{solve_each_view}
\begin{split}
&\mathop {\min }\limits_{\bm{\psi}(\bm{X}^v)} {tr\left( {{\bm{\psi}(\bm{X}^v)}{{(\bm{I}_N - {\bm{M}^v})}^T}(\bm{I}_N - {\bm{M}^v}){\bm{\psi}(\bm{X}^v)}^{^T}} \right)}\\ &+\gamma {\bm{w}^v tr\left( \bm{\psi}(\bm{X}^v) (\bm{D}^*-\bm{K}^*) {\bm{\psi}(\bm{X}^v)}^T  \right)} \\
& s.t.\hspace{0.5em}{\bm{\psi}(\bm{X}^v)}{\bm{\psi}(\bm{X}^v)}^{^T} = \bm{I}_{d^v} \\
\end{split}
\end{equation}
where $\bm{K}^*$ stands for the similarity matrix of the centroid view, and $\bm{D}^*=diag(\bm{d}_{11}^*, \bm{d}_{22}^*, \ldots, \bm{d}_{NN}^*)$ is a diagonal matrix, $\bm{d}_{NN}^*=\sum_{i=1}^N{\bm{K}_{iN}^*}$.
Set $\bm{L}^v={(\bm{I}_N - {\bm{M}^v})}^T(\bm{I}_N - {\bm{M}^v})+\gamma{\bm{w}^v (\bm{D}^*-\bm{K}^*)}$. Therefore, with the constraint ${\bm{\psi}(\bm{X}^v)} {\bm{\psi}(\bm{X}^v)}^{^T} = \bm{I}_{d^v}$, the optimal $\bm{\psi}(\bm{X}^v)$ could be solved by eigen-decomposition. $\bm{\psi}(\bm{X}^v)$ consists of eigenvectors corresponding to the smallest $d^v$ eigenvalues. 

$\bm{Updating} \quad \bm{w}$: By fixing all variables but $\bm{w}^v$, we could calculating $\bm{w}^v$ by Eq.(\ref{solve_w}) and normalization for each view as follows:
\begin{equation}\label{solve_final_w}
\begin{split}
&\bm{w}^v=\frac{{tr\left( \bm{U}^* (\bm{D}^v-\bm{K}^v) {\bm{U}^*}^T \right)}^{\frac{p}{2}-1}}{\sum_{v=1}^m{{tr\left(\bm{U}^* (\bm{D}^v-\bm{K}^v) {\bm{U}^*}^T \right)}^{\frac{p}{2}-1}}}\\
\end{split}
\end{equation}
It's notable that the value of $p$ could directly influence the weighting factor $\bm{w}$. When $p \to 0$, $\bm{w}^v$ is proportional to the reciprocal of the disagreement term between the $v$th view and centroid view. Conversely, when $p \to 2$, all elements in $\bm{w}$ tend to be equal to $1/m$. 

\begin{algorithm}
\caption{The optimization procedure of MVLPE}
\label{algorithm}
\hspace*{0.02in} {\bf Require:}

\hspace*{0.05in}1. A multi-view features set with N training samples having m views ${\bm{X}^v} = [\bm{x}_1^v,\bm{x}_2^v, \ldots ,\bm{x}_N^v] \in {\mathbb{R}^{{D_v} \times N}}, v=1,2,\dots,m$.

\hspace*{0.05in}2. Set the parameters $\gamma$ and $p$ in Eq.(\ref{MvLPE}).

\hspace*{0.02in} {\bf The Main Procedure:}
\begin{algorithmic}
\FOR{\emph{v}=1:\emph{m}}
    \STATE{3. Initialize $\bm{w}^v=1/m$.}
    \STATE{4. Specialize the $\bm{M}^v$ in Eq.(\ref{low_rank_coef}).}
    \STATE{5. Initialize $\bm{\psi}(\bm{X}^v)$ according to $M^v$}
\ENDFOR

\REPEAT
\STATE{6. Update $\bm{U}^*$ by solving Eq.(\ref{solve_common_view}).}
\FOR{\emph{v}=1:\emph{m}}
    \STATE{7. Update $\bm{\psi}(\bm{X}^v)$ for the $v$th view by solving Eq.(\ref{solve_each_view}).}
\ENDFOR
\STATE{8. Update $\bm{w}$ by solving Eq.(\ref{solve_w}).}
\UNTIL{$[\bm{U}^*, \bm{\psi}(\bm{X}^1),\bm{\psi}(\bm{X}^2), \ldots, \bm{\psi}(\bm{X}^m)]$ converges.}

{\bf return }\hspace*{0.05in} $\bm{U}^*$.
\end{algorithmic}
\end{algorithm}

\subsection{Convergence Analysis}
Because our proposed MvLPE is solved by alternating optimization strategy, it's essential to analyze its convergence. We first need to utilize the following lemma introduced by \cite{nie2012low}.

\textbf{Lemma 1.} For any positive number a and b, the following inequality holds:
\begin{equation}
\begin{split}
& a^{\frac{p}{2}}-\frac{p}{2}\frac{a}{b^{1-\frac{p}{2}}} \le b^{\frac{p}{2}}-\frac{p}{2}\frac{b}{b^{1-\frac{p}{2}}}\\
\end{split}
\end{equation}

\textbf{Theorem 1.} Each updated $\bm{U}^*$ in \textbf{Algorithm \ref{algorithm}} will monotonically decreases the objective in Eq.(\ref{MvLPE_0}) in each iteration.

\textbf{Proof:} We use $\widetilde{\bm{U}}^*$ to denote the updated $\bm{U}^*$ in each iteration. According to the optimization to $\bm{U}^*$ in \textbf{Algorithm \ref{algorithm}}, we know that $\widetilde{\bm{U}}^*$ makes the objective of Eq.(\ref{solve_common_view}) have the smaller than $\bm{U}^*$. Combining $\bm{w}$ computed in \textbf{Algorithm \ref{algorithm}}, we could drive:
\begin{equation}\label{ineq1}
    \begin{split}
    &
    \sum\limits_{v = 1}^m \frac{p}{2} {\frac{tr\left( {\widetilde{\bm{U}}^*}(\bm{D}^v-\bm{K}^v) \widetilde{\bm{U}}^{*^T} \right)}{{tr\left( \bm{U}^* (\bm{D}^v-\bm{K}^v) \bm{U}^{*^T} \right)}^{1-\frac{p}{2}}}}\\
    &\le
        \sum\limits_{v = 1}^m \frac{p}{2} {\frac{tr\left( {\bm{U}^*}^T(\bm{D}^v-\bm{K}^v) \right)}{{tr\left( \bm{U}^* (\bm{D}^v-\bm{K}^v) \bm{U}^{*^T} \right)}^{1-\frac{p}{2}}}} \\
    \end{split}
\end{equation}
According to \textbf{Lemma 1}, we have:
\begin{scriptsize}
\begin{equation}\label{ineq2}
    \begin{split}
        &{\sum\limits_{v = 1}^m {tr\left( \widetilde{\bm{U}^*}(\bm{D}^v-\bm{K}^v) \widetilde{\bm{U}}^{*^T}  \right)}}^{\frac{p}{2}}-\sum\limits_{v = 1}^m \frac{p}{2} {\frac{tr\left( {\widetilde{\bm{U}}^*}(\bm{D}^v-\bm{K}^v) \widetilde{\bm{U}}^{*^T} \right)}{{tr\left( \bm{U}^* (\bm{D}^v-\bm{K}^v) \bm{U}^{*^T} \right)}^{1-\frac{p}{2}}}}\\
        &\le
        {\sum\limits_{v = 1}^m {tr\left( \bm{U}^*(\bm{D}^v-\bm{K}^v) \bm{U}^{*^T}  \right)}}^{\frac{p}{2}}-\sum\limits_{v = 1}^m \frac{p}{2} {\frac{tr\left( {\bm{U}^*}^T(\bm{D}^v-\bm{K}^v) \right)}{{tr\left( \bm{U}^* (\bm{D}^v-\bm{K}^v) \bm{U}^{*^T} \right)}^{1-\frac{p}{2}}}}\\
    \end{split}
\end{equation}
\end{scriptsize}

Sum over Eq.(\ref{ineq1}) and Eq.(\ref{ineq2}) in the two sides, we could derive:
\begin{small}
\begin{equation}
    \begin{split}
        {\sum\limits_{v = 1}^m {tr\left( \widetilde{\bm{U}^*}(\bm{D}^v-\bm{K}^v) {\widetilde{\bm{U}^*}}^T  \right)}}^{\frac{p}{2}}
        \le
        {\sum\limits_{v = 1}^m {tr\left( \bm{U}^*(\bm{D}^v-\bm{K}^v) {\bm{U}^*}^T  \right)}}^{\frac{p}{2}}
    \end{split}
\end{equation}
\end{small}

Thus the alternating optimization will monotonically decrease the objective in Eq.(\ref{MvLPE_0}). 

\textbf{Theorem 2.} The objective function in Eq.(\ref{MvLPE}) is bounded. The proposed optimization algorithm monotonically decreases the loss value in each step, which makes the solution convergence to a local optimum.

\textbf{Proof:} It's easy to find that there must exist one view which can make $e_{min}=tr\left( {\bm{U}^v{{({\bm{I}_N} - \bm{M}^v)}^T} ({\bm{I}_N} - \bm{M}^v){{\bm{U}^v}^T}} \right)>0$ to be smallest among all views. Similarly, we also find such a view that is closest to the centroid view, that is $d_{min}>0$. Because the hyperparameter $\gamma>0$, it is provable that the objective value in Eq.(\ref{MvLPE_0}) is greater than $m(e_{min}+d_{min})$. Therefore, The objective function in Eq.(\ref{MvLPE}) has a lower bound.

For \textbf{Algorithm 1}, it's obvious to see that \{${\bm{\psi}(\bm{X}^1)},{\bm{\psi}(\bm{X}^2)}, \ldots ,{\bm{\psi}(\bm{X}^m)}$\} generated via solving Eq.(\ref{solve_each_view}) are the exact minimum points of Eq.(\ref{solve_each_view}) respectively. As a result, the value of the objective function on \{${\bm{\psi}(\bm{X}^1)},{\bm{\psi}(\bm{X}^2)}, \ldots ,{\bm{\psi}(\bm{X}^m)}$\} in Eq.(\ref{MvLPE}) is decreasing in each iteration of \textbf{Algorithm 1}. Combining this with \textbf{Theorem 1}, thus the alternating optimization will monotonically decrease the objective in Eq.(\ref{MvLPE}). Therefore, according to the bounded monotone convergence theorem \cite{rudin1964principles} that asserts the convergence of every bounded monotone sequence, the proposed optimization algorithm converges.

\subsection{Extensions}
Differing from these multi-views methods existing limitations of generalization and scalability, we could extend those single view-based methods, which could be cast as a special form of the quadratically constrained quadratic program (QCQP), into multi-view setting referring to MvLPE. Specially, such methods \cite{belkin2002laplacian, wold1987principal, mika1999fisher, he2005neighborhood, he2004locality, qiao2010sparsity, xu2007marginal} could be solved by the following equation:
 \begin{equation}
 \label{single view}
\begin{split}
&\mathop {\min }\limits_{{\bm{U}^v}} tr(\bm{U}^v \bm{Q}^v \bm{U}^{v^T}) \\
& s.t.  \quad {\bm{U}^v}{\bm{C}^v}{\bm{U}^v}^{^T} = \bm{I} \\
\end{split}
\end{equation}
 where $\bm{M}^v \in {\mathbb{R}^{N \times N}}$ reflects the intrinsic structure for the $v$th view and $\bm{C}^v$ stands for the different constraint term according to different methods. For the example of LPE, we could utilize $\bm{I}_N$ to reformulate ${\bm{C}^v}$. Taking LDA \cite{mika1999fisher}, NPE \cite{he2005neighborhood}, and LPP \cite{he2004locality} as examples, we could express $\bm{M}^v$ and $\bm{C}^v$ as follows:
\begin{itemize}
    
    \item \textbf{LDA:} $\bm{Q}_{i,j}^v=-1/N_v^c$ if $\bm{X}_i^v$ and $\bm{X}_j^v$ belong to the class $c$, 0 otherwise, where $N_v^c$ is the number of samples for class $c$ in the $v$th view. And $\bm{C}^v=\bm{M}^v-\bm{I}_N$, where $\bm{I}_N \in \mathbb{R}^{N \times N}$ is an identity matrix.
    
    \item \textbf{NPE:} $\bm{Q}^v={({\bm{I}_N-\bm{S}^v})}^T(\bm{I}_N-\bm{S}^v)$, where $\bm{S}^v \in \mathbb{R}^{N \times N}$ is the reconstruction coefficient matrix in the $v$th view. And $\bm{C}^v=\bm{I}_N$, where $\bm{I}_N \in \mathbb{R}^{N \times N}$ is an identity matrix.
    
    \item \textbf{LPP:} $\bm{Q}^v$ is the Laplacian matrix in the $v$th view and $\bm{C}^v$ is a diagonal matrix, in which $\bm{C}_{ii}^v$ is the sum of all elements in the $i$th row of $\bm{Q}^v$. 
\end{itemize}

To further improve the performance of those single-view representation methods, we could also provide with three manners, including direct embedding, linear projection, and kernel tricks, to keep their intrinsic information as much as possible. Then, we could extend such QCQP-specific single-view methods into multi-view setting to integrate the information from multiple views according to the construction process of the proposed MvLPE. As a result, we could take full advantage of these works based on single view meanwhile integrating rich information among different views.

\section{Experiments}

In this section, we evaluate the performance of MvLPE compared to several classical single-view and multi-view learning methods in the multi-view datasets of texts and images. We first introduce the details of the utilized datasets and comparing methods in \ref{expereient1}. Then we show the experiments in \ref{expereient2} and \ref{expereient3}. These experiment results verify the excellent performance of MvLPE. Finally, we empirically validate the convergence of MvLPE in \ref{expereient4} according to the curve of objectives.

\subsection{Datasets and Competitors}\label{expereient1}

Texts and images are usually represented by multi-view features, and the feature in each view is represented in high-dimensional space. Therefore, we conduct our experiments on six datasets in the form of texts and images. Three text datasets adopted in the experiments are widely used in works, including WebKB\footnotetext[1]{http://www.webkb.org/}, 3Source\footnotetext[2]{http://mlg.ucd.ie/datasets/3sources.html}, Cora\footnotetext[3]{http://lig-membres.imag.fr/grimal/data.html}. Three images datasets adopted in the experiments are widely used in works, including: ORL\footnotetext[4]{http://www.uk.research.att.com/facedatabase.html}, Yale\footnotetext[5]{http://cvc.yale.edu/projects/yalefaces/yalefaces.html}, Caltech101\footnotetext[6]{http://www.vision.caltech.edu/ImageDatasets/Caltech101/}. We extract features for images using three different image descriptors including LBP, Gist, and EDH. To be more specific, those datasets are summarized as follows:

\textbf{WebKB} contains 4 subsets of documents over six labels and each subset consists of three views, including the text, the anchor text, and the title. 

\textbf{3Sources} consists of 3 well-known online news sources: BBC, Reuters, and the Guardian, and each source is treated as one view. We select the 169 stories which are reported in all these 3 sources.

\textbf{Cora} consists of 2708 scientific publications which come from 7 classes. Because the document is represented by content and citation views, Cora could be considered as a two-views dataset. 

\textbf{ORL} and \textbf{Yale} are two face image datasets that have been widely used in face recognition, where ORL consists of 400 faces corresponding to 40 peoples and Yale consists of 165 faces from 15 peoples.

\textbf{Caltech101} is a benchmark image dataset that contains 9144 images corresponding to 102 objects and it's a benchmark dataset for image classification. 

\begin{table}[htbp]
\caption{The detail information of the multi-view datasets}
\label{datasets}
\centering
\begin{tabular*}{0.45\textwidth}{@{\extracolsep{\fill}}cccc}  
\hline
Datasets &Samples &Classes &Views\\
\hline  
WebKB & 226 & 4 & Text, Anchor Text, and Title\\
3Sources & 169 & 6 & BBC, Reuters, and Guardian\\
Cora & 2708 & 7 & Content, and Cites\\
ORL &400 & 40 & LBP, Gist, and EDH\\
Yale &165 & 15 & LBP, Gist, and EDH\\
Caltech101 & 9144 & 102 & LBP, Gist, and EDH\\
\hline
\end{tabular*}
\end{table}

More specifically, all views information of these utilized datasets is summarized in Table \ref{datasets}. The effectiveness of MvLPE is evaluated by comparing it with the following algorithms, including the best performance of the single view based LE(BLE), the feature concatenation based LE(CLE), MSE, Auto-weighted \cite{nie2017auto}, Co-regularized, Co-training \cite{kumar2011co2}, MvCCA. Besides, we also compared the single view low-dimensional embedding in our framework with original low-dimensional embedding using MvLPE, and additional experiments on the single feature in multi-view framework by correcting and complemented by ones from other views are to verify the fact that our method could make use of complementary information among different views by correcting and complementing ones from other views.

\subsection{Experiments on textual datasets}\label{expereient2}
To show the superior performance of MvLPE, the experiments on three multi-view textual datasets (WebKB, 3Source, and Cora) are conducted in this section. And 1NN classifier is adopted here to classify all testing samples to verify the performances of all methods when we have obtained the low-dimensional embedding using all methods. And the mean(MEAN) and max(MAX) classification accuracy on multi-view datasets are employed as the evaluation index.

For WebKB dataset, we randomly select 50\% of the samples for each subset as training samples every time. The embedding dimensionality of all the methods is set as 30. We run all methods 20 times with different random training samples and testing samples. Table \ref{WebKB} shows the MEAN and MAX value on WebKB dataset.

\begin{table*}[htbp]
\caption{The classification accuracy on 3Source dataset}
\label{WebKB}
\centering  %
\begin{tabular*}{0.95\textwidth}{@{\extracolsep{\fill}}lllllllll}  
\hline
Methods & \multicolumn{2}{c}{WebKB-1} & \multicolumn{2}{c}{WebKB-2} & \multicolumn{2}{c}{WebKB-3} & \multicolumn{2}{c}{WebKB-4} \\
\cline{2-3} \cline{4-5} \cline{6-7} \cline{8-9}
 & MEAN(\%) & MAX(\%) & MEAN(\%) & MAX(\%)& MEAN(\%) & MAX(\%) & MEAN(\%) & MAX(\%)\\
\hline
BLE & 78.67 & 84.07 & 71.56 & 77.77 & 66.16 & 74.40 & 74.52 & 79.22 \\
CLE & 67.48 & 75.39 & 71.79 & 78.90 & 70.44 & 75.78 & 76.40 & 82.46 \\
MSE & 81.06 & 86.72 & \textbf{82.64} & \textbf{87.10} & 82.10 & 89.50 & 80.46 & 85.15\\
Auto-weighted & 82.18 & 84.11 & 78.91 & 84.12 & 81.43 & 87.50 & 72.79 & 82.64 \\
Co-regularized & 81.05 & 88.64 & 73.12 & 80.95 & 80.30 & 85.71 & 80.12 & 84.41 \\
Co-training & 81.94 & 90.17 & 73.01 & 78.89 & 77.39 & 82.03 & 79.66 & 84.24 \\
MvCCA & 82.17 & 89.64 & 78.71 & 81.58 & 77.78 & 79.12 & 72.73 & 79.85 \\
MvLPE & \textbf{83.17} & \textbf{91.84} & 78.86 & 84.21 & \textbf{85.28} & \textbf{92.96} & \textbf{79.96} & \textbf{85.63} \\
\hline
\end{tabular*}
\end{table*}

For 3Source dataset, we randomly select 50\% of the samples as training samples and remaining samples as testing samples every time. The dimension of the embedding obtained by all methods all maintains 20 and 30 dimensions. We run all methods 20 times with different random training samples and testing samples. Table \ref{3Source} shows the MEAN and MAX value on 3Source dataset.

\begin{table}[!htb]
\caption{The classification accuracy on 3Source dataset}
\label{3Source}
\centering  %
\begin{tabular}{lllll}  
\hline
Methods & \multicolumn{2}{c}{Dims=20} & \multicolumn{2}{c}{Dims=30} \\
 & MEAN(\%) & MAX(\%) & MEAN(\%) & MAX(\%)\\
\hline
BLE & 66.47 & 74.11 & 59.72 & 69.41 \\
CLE & 66.50 & 74.71 & 62.78 & 72.94 \\
MSE & 50.47 & 57.64 & 46.86 & 60.00 \\
Auto-weighted & 49.92 & 57.64 & 48.15 & 56.47 \\
Co-regularized  & 81.25 & 87.05 & 78.50 & 85.88\\
Co-training & 80.80 & 88.23 & \textbf{80.37} & 90.58  \\
MvCCA & 53.88 & 76.45 & 54.37 & 73.56 \\
MvLPE & \textbf{82.64} & \textbf{89.41} & 79.70 & \textbf{90.9} \\
\hline
\end{tabular}
\end{table}

For Cora dataset, we randomly select 50\% of the samples as training samples and remaining samples as testing samples every time. The dimension of the embedding obtained by all methods all maintains 20 and 30 dimensions. We run all methods 20 times with different random training samples and testing samples. Table \ref{Cora} shows the MEAN and MAX value on Cora dataset.

\begin{table}[!htb]
\caption{The classification accuracy on Cora dataset}
\label{Cora}
\centering  %
\begin{tabular}{lllll}  
\hline
Methods & \multicolumn{2}{c}{Dims=20} & \multicolumn{2}{c}{Dims=30} \\
 & MEAN(\%) & MAX(\%) & MEAN(\%) & MAX(\%)\\
\hline
BLE & 58.98 & 60.85 & 61.05 & 63.44 \\
CLE & 51.00 & 53.61 & 52.86 & 55.31 \\
MSE & 64.65 & 66.24 & 67.72 & 69.64 \\
Auto-weighted & 63.71 & 65.73 & 66.90 & 69.57 \\
Co-regularized  & 55.73 & 57.45 & 57.19 & 59.01 \\
Co-training & 70.53 & 72.23 & 72.11 & 73.54  \\
MvCCA & 71.11 & 72.35 & 71.52 & 72.05 \\
MvLPE & \textbf{73.7} & \textbf{75.23} & \textbf{73.45} & \textbf{75.84} \\
\hline
\end{tabular}
\end{table}

\subsection{Experiments on images datasets}\label{expereient3}
To show the superior performance of our framework, the experiments on three multi-view images datasets (Yale, ORL, Caltech101) are conducted in this section. And 1NN classifier is adopted here to classify all testing samples to verify the performances of all methods when we have obtained the embedding using all methods.

\begin{figure}[htbp]
\centering
\includegraphics[width=0.45\textwidth]{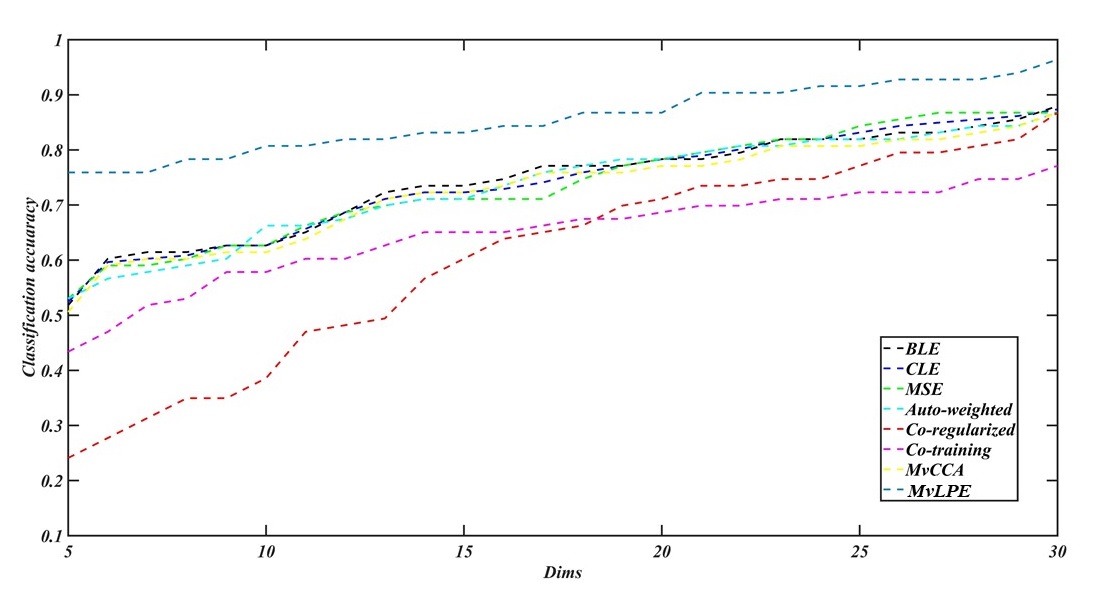}
\caption{The classification accuracy on Yale dataset}
\label{Yale results}
\end{figure}

For Yale dataset, we extract gray-scale intensity, local binary patterns, and edge direction histogram as 3 views. The dimension of embedding obtained by all methods all maintains from 5 to 30 dimensions. we randomly select
50\% samples as training ones while the other samples are assigned as the testing ones every time and run all methods 30 times with different random training samples and testing samples. Fig. \ref{Yale results} shows the accuracy values on Yale dataset.

\begin{figure}[htbp]
\centering
\includegraphics[width=0.45\textwidth]{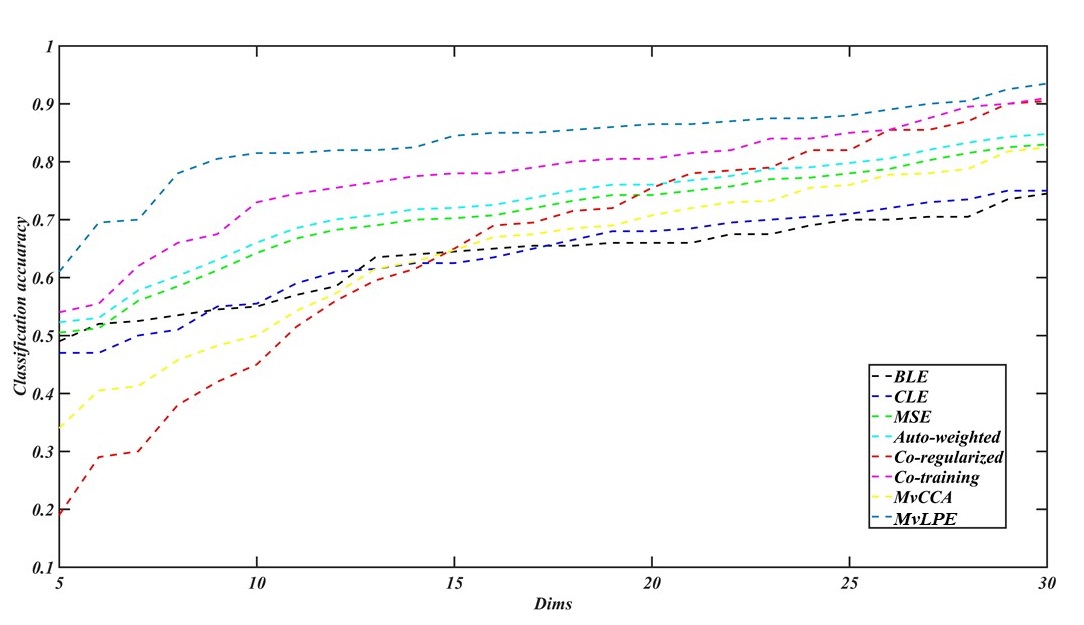}
\caption{The classification accuracy on ORL dataset}
\label{ORL results}
\end{figure}

For ORL dataset, we also randomly select 50\% samples as training ones while the other samples are assigned as the testing ones every time and run all methods 30 times with different random training samples and testing samples. And gray-scale intensity, local binary patterns, and edge direction histogram are utilized as 3 views. The dimension of embedding obtained by all methods all maintains from 5 to 30 dimensions. Fig. \ref{ORL results} shows the accuracy values on ORL dataset.

\begin{figure*}[htbp]
\centering
\subfigure[Dim=20]{
\centering
\includegraphics[width=0.45\textwidth]{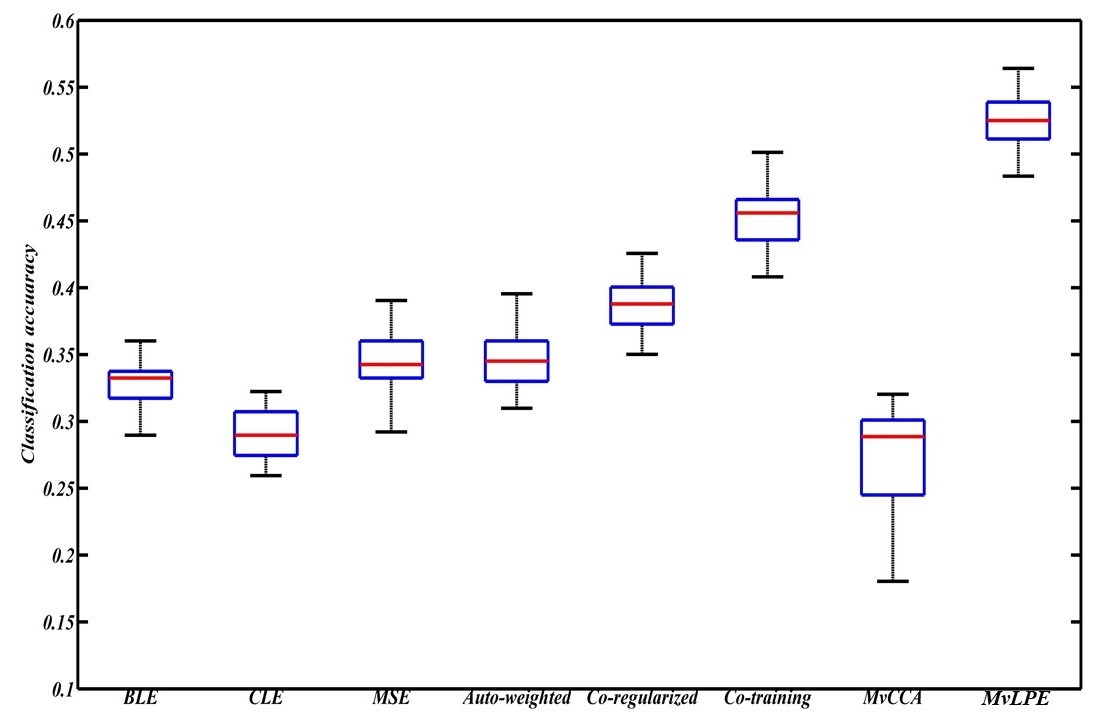}
}%
\subfigure[Dim=30]{
\centering
\includegraphics[width=0.45\textwidth]{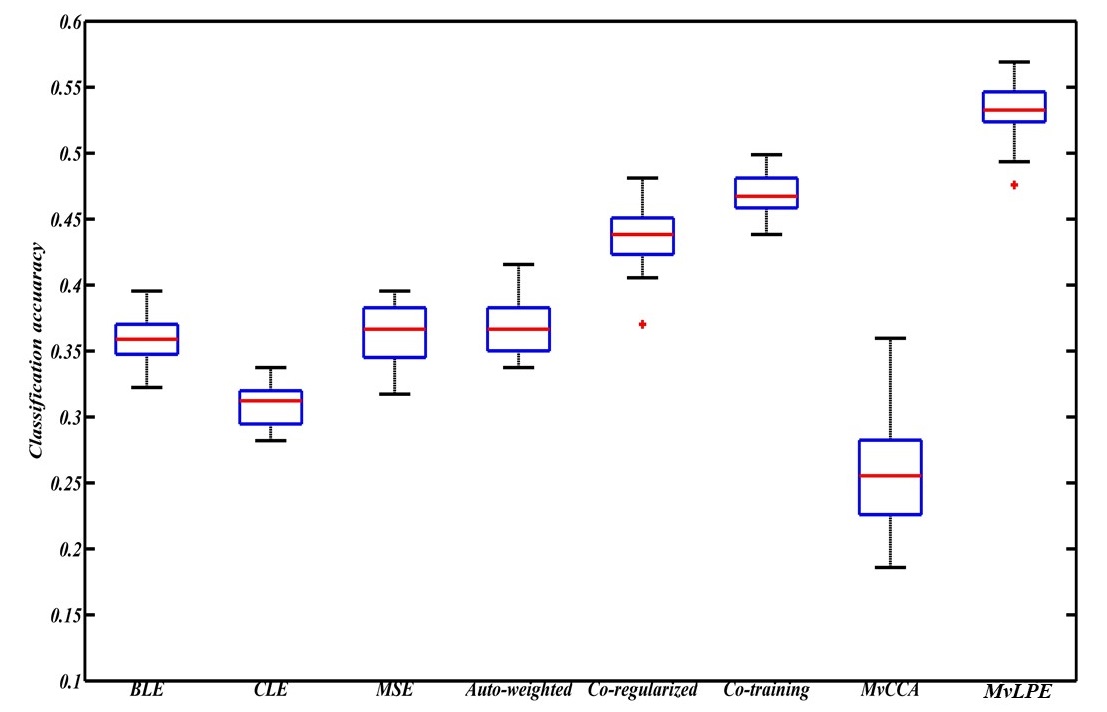}
}%
\caption{Classification results on Caltech101 dataset in different dimension}
\label{Caltech results}
\end{figure*}

For Caltech101 dataset, the first 20 classes are utilized in our experiments. Meanwhile, we extract EDH, LBP, and Gist features for an image as 3 views. The dimension of embedding obtained by all methods maintains 20 and 30 dimensions. We randomly select 50\% of the samples for Caltech101 dataset as training samples every time and run all methods 30 times with different random training samples and testing samples. Fig. \ref{Caltech results} shows the mean accuracy values on Caltech101 dataset.

\subsection{Convergence}\label{expereient4}
\begin{figure*}[htbp]
\centering
\subfigure[DIM=20 on Cora dataset]{
\centering
\includegraphics[width=0.45\textwidth]{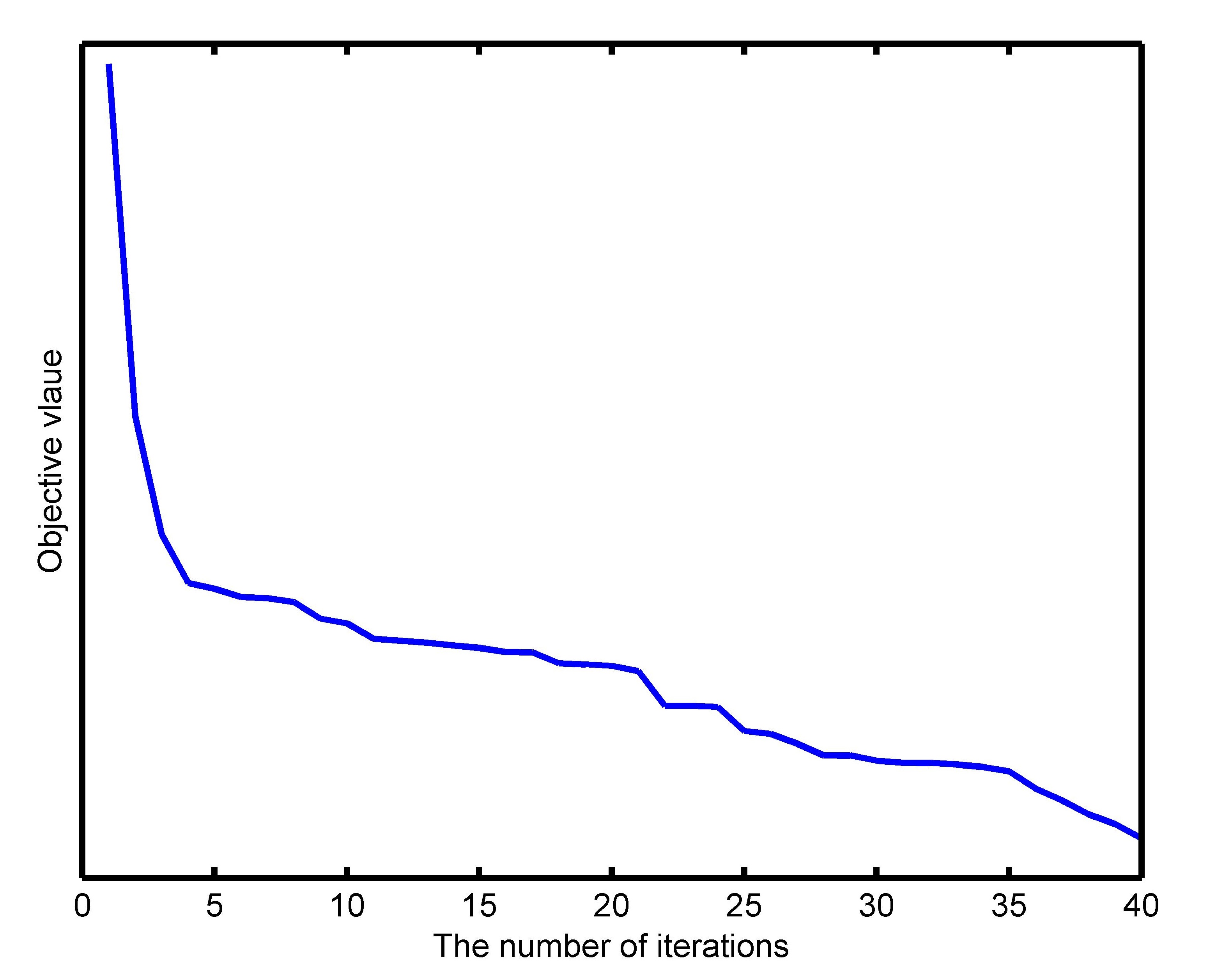}
}
\centering
\subfigure[DIM=30 on Cora dataset]{
\centering
\includegraphics[width=0.45\textwidth]{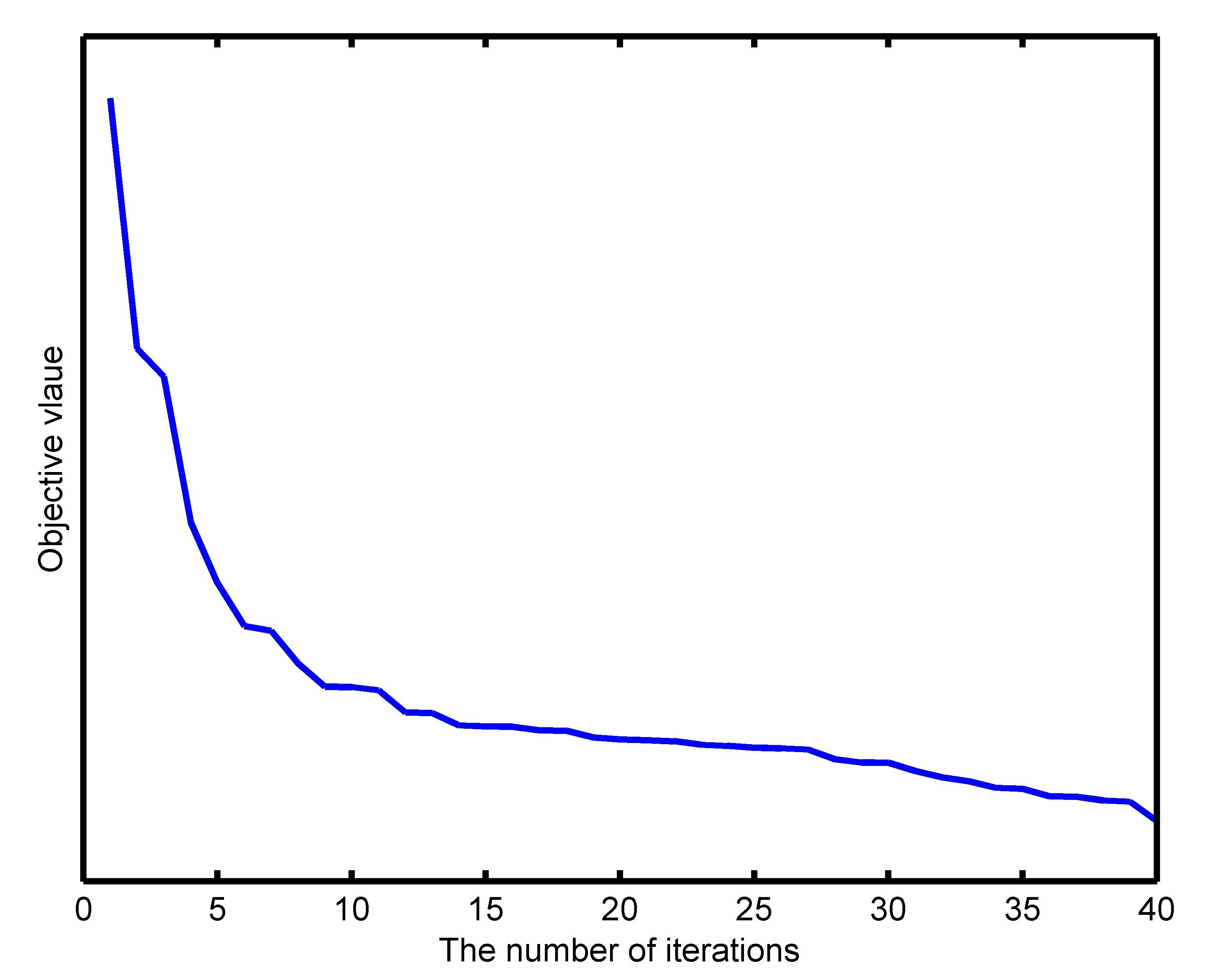}
}

\subfigure[DIM=20 on Caltech101 dataset]{
\centering
\includegraphics[width=0.45\textwidth]{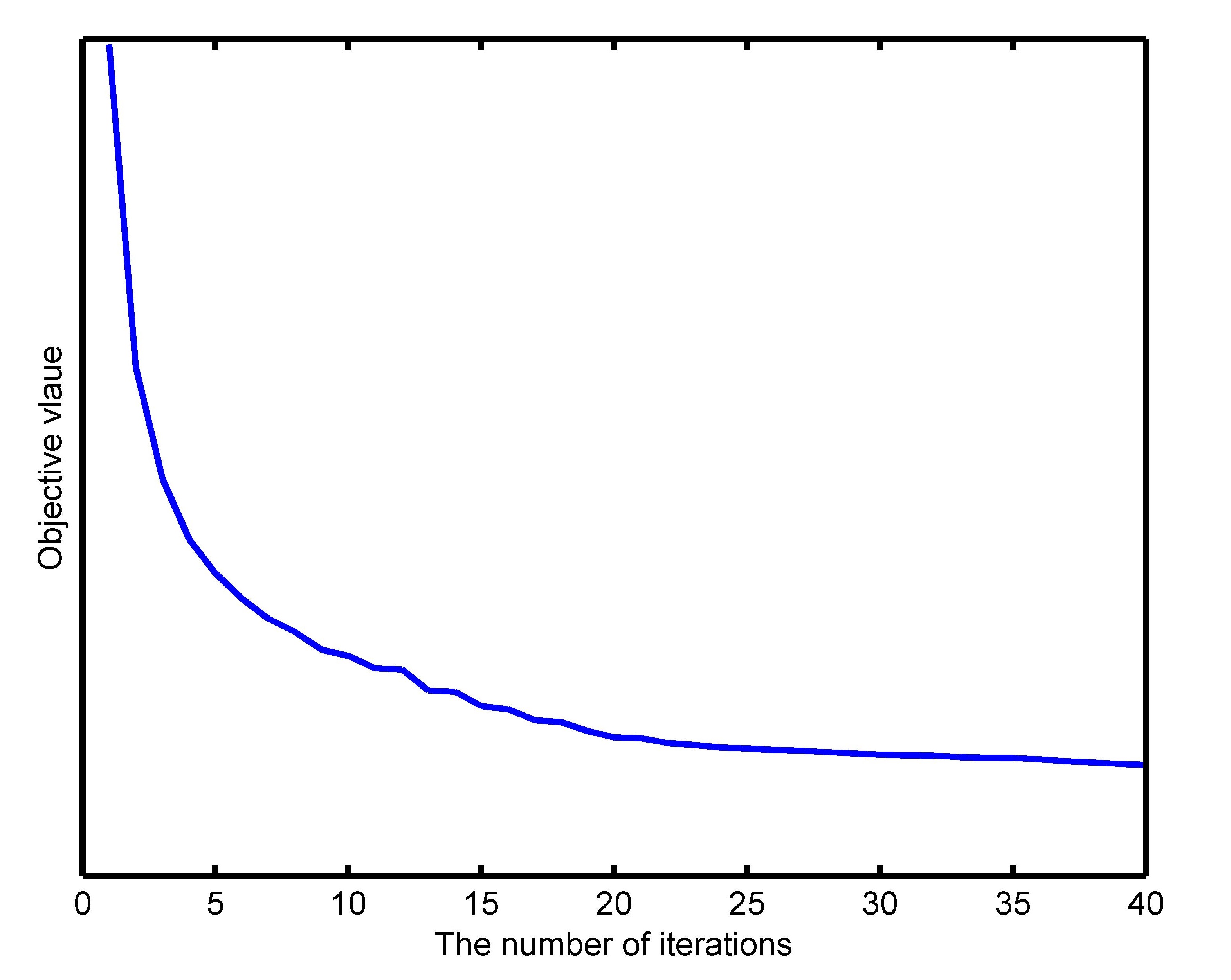}
}
\subfigure[DIM=30 on Caltech101 dataset]{
\centering
\includegraphics[width=0.45\textwidth]{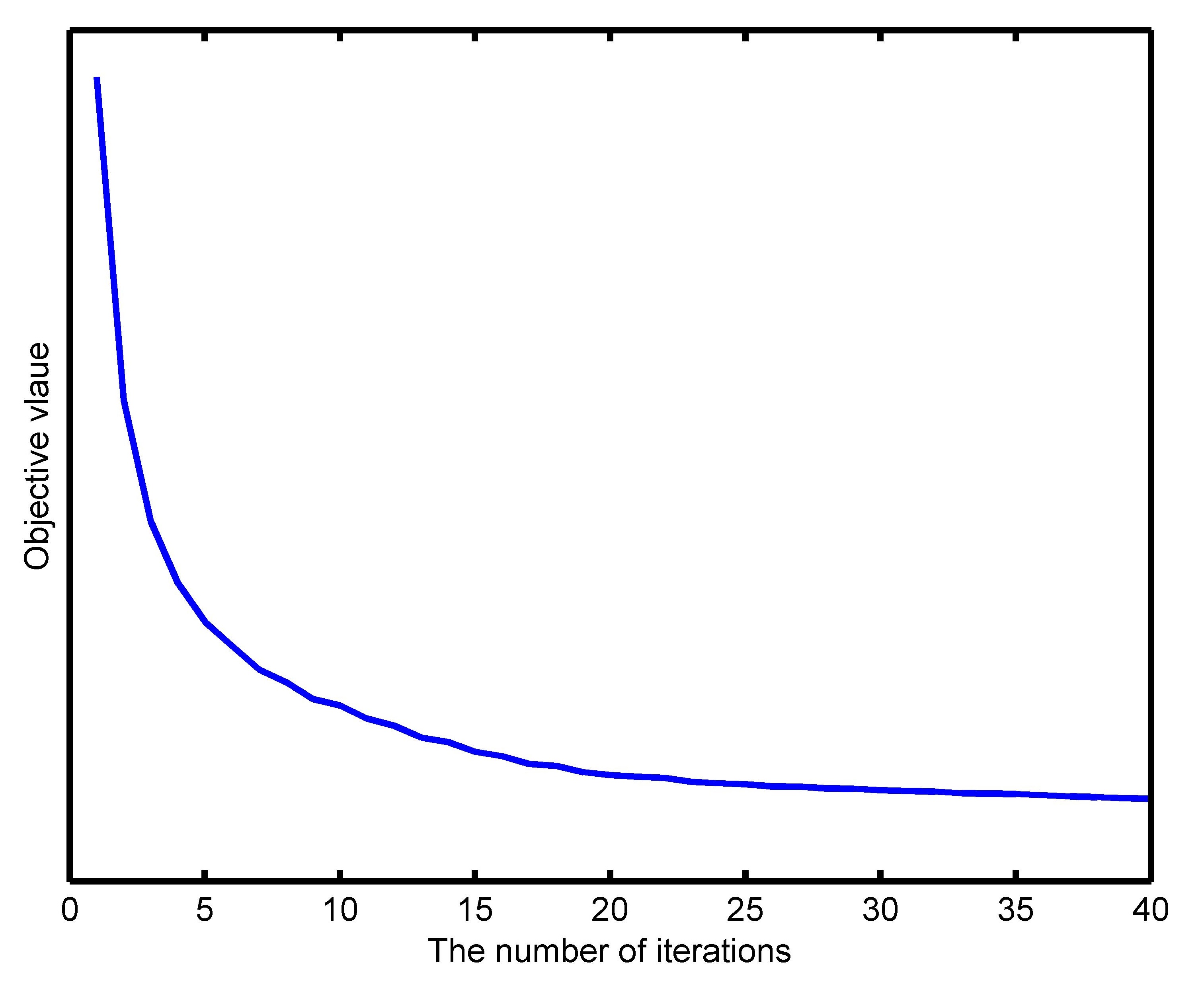}
}

\caption{Objective values of CMLLE on Cora and Caltech101}
\label{convergence}
\end{figure*}

Because our framework adopts an iterative procedure to obtain the optimal solution, it is essential to discuss the convergence in detail. In this section, we summarize the objective values of MvLPE on Cora and Caltech101 datasets according to the above experiments. All the training parameters (such as training numbers, dimensions) can be found above Fig.\ref{convergence}, which summarizes the objective values of Cora and Caltech101 datasets.

\section{Conclusion}
In this paper, we propose a novel multi-view learning method for multi-view representation, named Multi-view Low-rank Preserving Embedding (MvLPE). MvLPE deals with multi-view problems by integrating different views into one centroid view, which fully integrates compatible and complementary information from multi-view features set meanwhile maintaining low-rank reconstruction relations among samples for each view. Then, an iterative alternating strategy is adopted to find the optimal solution for our method and the optimization procedure is illustrated in detail. Moreover, we provide the convergence discussion of this method and its extensions for those single-view methods. The experiments have verified that the proposed MvLPE could effectively explore the underlying complementary information among multi-view data and achieve the superiority than other multi-view methods used in the experiments.

\section*{Acknowledgment}
This work was supported by National Natural Science Foundation of PR China(61672130, 61972064) and LiaoNing Revitalization Talents Program(XLYC1806006).

\bibliographystyle{IEEEtran}
\bibliography{IEEEexample}
\vspace{12pt}

\begin{IEEEbiography}[{\includegraphics[width=1in,height=1.25in,clip,keepaspectratio]{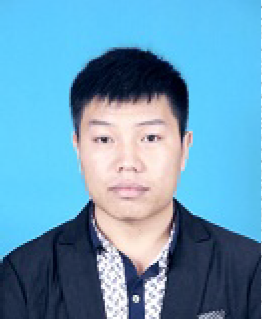}}]{Xiangzhu Meng}
received his BS degree from Anhui University, in 2015. Now he is working towards the PHD degree in School of Computer Science and Technology, Dalian University of Technology, China. His research interests include multi-view learning, deep learning and computing vision.
\end{IEEEbiography}

\begin{IEEEbiography}[{\includegraphics[width=1in,height=1.25in,clip,keepaspectratio]{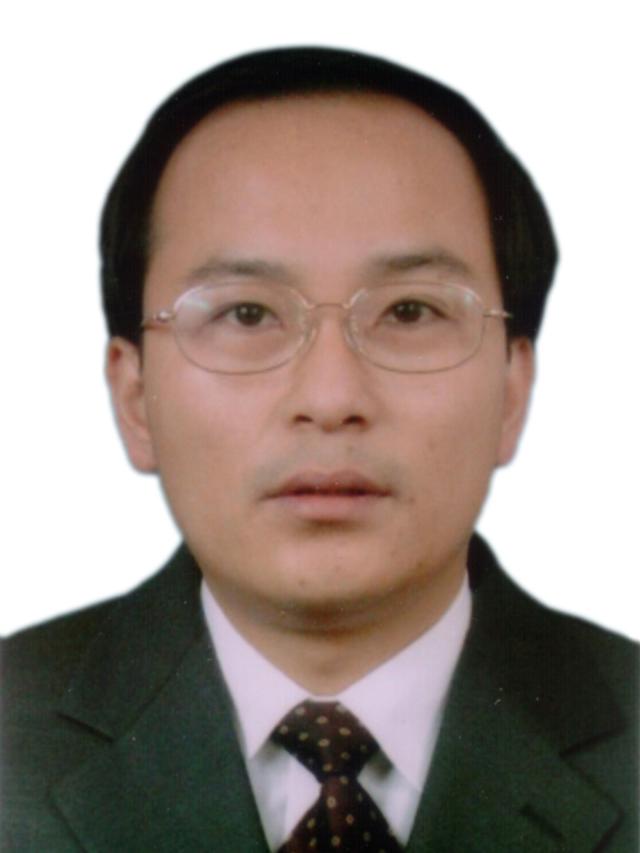}}]{Lin Feng}
received the BS degree in electronic technology from Dalian University of Technology, China, in 1992, the MS degree in power engineering from Dalian University of Technology, China, in 1995, and the PhD degree in mechanical design and theory from Dalian University of Technology, China, in 2004. He is currently a professor and doctoral supervisor in the School of Innovation Experiment, Dalian University of Technology, China. His research interests include intelligent image processing, robotics, data mining, and embedded systems.
\end{IEEEbiography}

\begin{IEEEbiography}[{\includegraphics[width=1in,height=1.25in,clip,keepaspectratio]{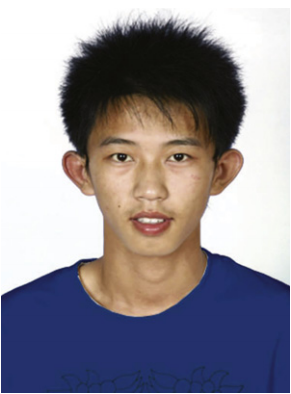}}]{Huibing Wang}
Huibing Wang received the Ph.D. degree in the School of Computer Science and Technology, Dalian University of Technology, Dalian, in 2018. During 2016 and 2017, he is a visiting scholar at the University of Adelaide, Adelaide, Australia. Now, he is a postdoctor in Dalian Maritime University, Dalian, Liaoning, China. He has authored and co-authored more than 20 papers in some famous journals or conferences, including TMM, TITS, TSMCS, ECCV, etc.
Furthermore, he serves as reviewers for TNNLS, Neurocomputing, PR Letters and MTAP, etc. His research interests include computing vision and machine learning

\end{IEEEbiography}

\end{document}